\documentstyle[aclap]{article}

\include{epsf}         

\title{\vspace{-0.5in}A Flexible POS Tagger Using an Automatically 
Acquired Language Model\thanks
{This research has been partially funded by the Spanish Research Department
(CICYT) and inscribed as TIC96-1243-C03-02}
} 

\author{Llu\'{\i}s M\`arquez \\
 LSI - UPC\\
 c/ Jordi Girona 1-3\\
 08034 Barcelona. Catalonia\\
 {\tt lluism@lsi.upc.es}\And
Llu\'{\i}s Padr\'o \\
 LSI - UPC\\
 c/ Jordi Girona 1-3\\
 08034 Barcelona. Catalonia\\
 {\tt padro@lsi.upc.es}}

\begin{document}
\bibliographystyle{fullname}
\maketitle
\vspace{-0.5in}
\begin{abstract}
We present an algorithm that automatically learns context
constraints using statistical decision trees. We then use the acquired
constraints in a flexible POS tagger.
The tagger is able to use information of any degree:
n-grams, automatically learned context constraints, 
linguistically motivated manually written constraints, etc. The sources and
kinds of constraints are unrestricted, and the language model can be
easily extended, improving the results. The tagger has been tested and
evaluated on the WSJ corpus. 
\end{abstract}

\section{Introduction}
In NLP, it is necessary to model the language in a representation
suitable for the task to be performed.
The language models more commonly used are based on two main
approaches:
first, the linguistic approach, in which the model is written by a
linguist, generally in the form of rules or constraints \cite{Voutilainen95}. 
Second, the automatic approach, in which the model is automatically
obtained from corpora (either raw or annotated)\footnote{When the model is obtained
from annotated corpora we talk about supervised learning, when it is
obtained from raw corpora training is considered unsupervised.}, 
and consists of n--grams \cite{Garside87,Cutting92}, rules \cite{Hindle89} or neural nets
\cite{Schmid94}.
In the automatic approach we can distinguish two main trends: The
low--level data trend collects statistics from the training corpora in
the form of n--grams, probabilities, weights, etc. The high level data
trend acquires more sophisticated information, such as context rules,
constraints, or decision trees
\cite{Daelemans96,Marquez95,Samuelsson96}. The acquisition
methods range from supervised--inductive--learning--from--example algorithms \cite{Quinlan86,Aha91} to
genetic algorithm strategies \cite{Losee94}, through the
transformation--based error--driven algorithm used in \cite{Brill95}.
Still another possibility are the hybrid models, which try to join the
advantages of both approaches \cite{Voutilainen97}.
\smallskip

We present in this paper a hybrid approach that puts together both
trends in automatic approach and the linguistic approach.
We describe a POS tagger based on the work described in
\cite{Padro96}, that is able to use bi/trigram information, 
automatically learned context constraints and 
linguistically motivated manually written constraints. The sources and
kinds of constraints are unrestricted, and the language model can be
easily extended. The structure of the tagger is presented in
figure~1. 
\medskip

\smallskip\noindent\hfil\epsfbox{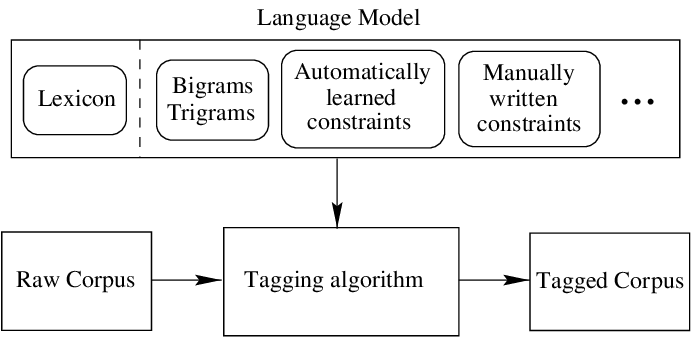}\hfil

\hfil \small{Figure~1: Tagger architecture.}\hfil
\bigskip
\normalsize

We also present a constraint--acquisition algorithm that uses
statistical decision trees to learn context constraints from annotated
corpora and we use the acquired constraints to feed the POS tagger.
\smallskip
  
The paper is organized as follows. 
In section \ref{seccio-model} we describe our language model, in
section \ref{seccio-regles} we describe the constraint acquisition
algorithm, and in section \ref{seccio-relax} we expose the tagging
algorithm. Descriptions of the corpus used, the experiments performed
and the results obtained can be found in sections \ref{seccio-corpus} 
and \ref{seccio-experiments}.

\section{Language Model}
\label{seccio-model}

We will use a hybrid language model consisting of
an automatically acquired part and a linguist--written part.
\smallskip

The automatically acquired part is divided in two kinds of
information: on the one hand, we have bigrams and trigrams collected
from the annotated training corpus (see section \ref{seccio-corpus}
for details). On the other hand, we have context constraints learned
from the same training corpus using statistical decision trees, as
described in section \ref{seccio-regles}.
\smallskip

The linguistic part is very small ---since there were no available 
resources to develop it further--- and covers only very few cases, but it
is included to illustrate the flexibility of the algorithm.

A sample rule of the linguistic part:
\begin{verbatim}
 10.0 (%vauxiliar%) 
    (-[VBN IN , : JJ JJS JJR])+ 
    <VBN>;
\end{verbatim}
 \indent This rule states that a tag {\em past participle} ({\bf VBN}) is very
 compatible (10.0) with a left context consisting of a {\bf \%vauxiliar\%} 
 (previously defined macro which includes all forms of ``have'' and ``be'') 
 provided that all the words in between don't have any of the tags in the set
 {\bf [VBN IN , : JJ JJS JJR]}. That is, this rule raises the support for
 the tag {\em past participle} when there is an auxiliary verb to the left but
 only if there is not another candidate to be a past participle or an 
 adjective inbetween. The tags {\bf [IN , :]} prevent the rule from being
 applied when the auxiliary verb and the participle are in two different
 phrases (a comma, a colon or a preposition are considered to mark the 
 beginning of another phrase).
 \smallskip

  The constraint language is able to express the same kind of patterns
than the Constraint Grammar formalism \cite{Karlsson95}, although in a
different formalism. In addition, each constraint has a compatibility 
value that indicates its strength. In the middle run, the system will 
be adapted to accept CGs.  

\section{Constraint Acquisition}
\label{seccio-regles}

Choosing, from a set of possible tags, the proper syntactic tag for a
word in a particular context can be seen as a problem of classification.
Decision trees, recently used in NLP basic tasks such as
tagging and parsing \cite{McCarthy95,Daelemans96,Magerman96}, are suitable
for performing this task.
\smallskip

A decision tree is a {\it n}--ary branching tree that represents a {\it classification rule} 
for classifying the {\it objects} of a certain domain into a set of
mutually exclusive {\it classes}.
The domain objects are described as a set of attribute--value pairs, 
where each {\it attribute} measures a relevant feature of an object
taking a (ideally small) set of discrete, mutually incompatible {\it values}.
Each non--terminal node of a decision tree represents  a question on
(usually) one attribute.  For each possible value of this attribute
there is a branch to follow. Leaf nodes represent concrete classes.

Classify a new object with a decision tree is simply 
following the convenient path through the tree until a leaf is reached.

{\it Statistical} decision trees only differs from common decision trees in that leaf
nodes define a conditional probability distribution on the set of classes.

It is important to note that decision trees can be directly translated 
to rules considering, for each path from the root to a leaf, the conjunction 
of all questions involved in this path as a condition and the class assigned to the
leaf as the consequence. Statistical decision trees would generate
rules in the same manner but assigning a certain degree of
probability to each answer. 

So the learning process of contextual constraints is performed 
by means of learning one statistical decision tree
for each class of POS ambiguity\footnote{
Classes of ambiguity are determined by the groups of possible tags 
for the words in the corpus, i.e,
{\it noun-adjective}, {\it noun-adjective-verb}, {\it
  preposition-adverb}, etc. }
and converting them to constraints (rules) expressing 
compatibility/incompatibility of concrete tags in certain contexts. 
\medskip

\noindent{\bf Learning Algorithm}
\smallskip

\noindent The algorithm we used for constructing the statistical decision trees
is a non--incremental supervised learning--from--examples algorithm 
of the TDIDT (Top Down Induction of Decision Trees) family.
It constructs the trees in a top--down way, guided by the distributional
information of the examples, but not on the examples order
\cite{Quinlan86}. 
Briefly, the algorithm works as a recursive process that departs from
considering the whole set of examples at the root level and constructs
the tree in a top--down way branching at any non--terminal node
according to a certain {\it selected} attribute. 
The different values of this attribute induce a partition of the
set of examples in the corresponding subsets, in which 
the process is applied recursively in order to generate the 
different subtrees. 
The recursion ends, in a certain node,  either when all (or almost all) the 
remaining examples belong to the same class, or when the number of examples 
is too small. 
These nodes are the leafs of the tree and contain the conditional
probability distribution, of its associated subset of examples, on the 
possible classes.

The heuristic function for selecting the most useful attribute at each
step is of a crucial importance in order to obtain simple trees, since
no backtracking is performed. There exist two main families of 
attribute--selecting functions: 
{\it information}--based~\cite{Quinlan86,Lopez91} and
{\it statistically}--based~\cite{Breiman84,Mingers89a}.
\medskip

\noindent{\it Training Set}
\smallskip

\noindent For each class of POS ambiguity the initial
example set is built by selecting from the training corpus
all the occurrences of the words belonging to this ambiguity class.
More particularly, the set of attributes that describe each example 
consists of the part--of--speech tags of the neighbour words, and the
information about the word itself (orthography and the proper tag in
its context). The window considered in the experiments reported in
section \ref{seccio-experiments} is 3 words to the left and 2 to the right.
The following are two real examples from the training set for the 
words that can be preposition and adverb at the same time
(IN--RB conflict).
\smallskip

\indent\indent{\tt VB DT NN  <"as",IN>  DT JJ}
\newline
\indent\indent{\tt NN IN NN  <"once",RB>  VBN TO}
\smallskip

Approximately $90\%$ of this set of examples is used for the
construction of the tree. The remaining $10\%$ is used as fresh
test corpus for the pruning process.
\medskip

\noindent{\it Attribute Selection Function}
\smallskip

\noindent For the experiments reported in
section \ref{seccio-experiments} we used a
attribute selection function due to L\'opez de M\'an\-ta\-ras \cite{Lopez91}, 
which belongs to the information--based family.
Roughly speaking, it defines a distance measure between partitions and
selects for branching the attribute that generates the closest
partition to the {\it correct partition}, namely the one that joins
together all the examples of the same class.

Let $X$ be a set of examples, ${\cal C}$ the set of classes
and $P_{\cal C}(X)$ the partition
of $X$ according to the values of ${\cal C}$.
The selected attribute will be the one 
that generates the closest partition of $X$ to $P_{\cal C}(X)$. 
For that we need to define a distance measure between partitions.
Let $P_A(X)$ be the partition of $X$ induced by the values of
attribute $A$.
The average information of such partition is defined as follows:
$$
I(P_A(X)) = {\displaystyle - \sum_{a\in P_A(X)} \; p(X,a) \,\log_2
  p(X,a)}\,,
$$
\noindent where $p(X,a)$ is the probability for an element of $X$ belonging to the set 
$a$ which is the subset of $X$ whose examples have a certain value for the attribute $A$,
and it is estimated by the ratio $\frac{|X\cap\; a|}{|X|}$. This average information 
measure reflects the randomness of distribution of the elements of $X$ between the classes of 
the partition induced by $A$. If we consider now the intersection between two different partitions 
induced by attributes $A$ and $B$ we obtain
\medskip

\noindent $I(P_A(X)\cap P_B(X)) =$
\smallskip

\hfill${\displaystyle - \sum_{a\in P_A(X)}\sum_{b\in P_B(X)} \; p(X,a\!\cap\! b) 
\,\log_2 p(X,a\!\cap\! b)}\,$.
\medskip

\noindent Conditioned information of $P_B(X)$ given $P_A(X)$ is
\medskip

\noindent $I(P_B(X)| P_A(X))=$
\smallskip

\indent $I(P_A(X)\cap P_B(X)) - I(P_A(X))=$
\smallskip

\hfill${\displaystyle -\!\sum_{a\in P_A(X)}\sum_{b\in P_B(X)} \; p(X,a
   \!\cap\! b) \,\log_2 {\frac {p(X,a\!\cap\! b)}{p(X,a)}}}\,$.
\medskip

\noindent It is easy to show that the measure
\medskip

\noindent $d(P_A(X),P_B(X)) =$
\smallskip

\hfill$I(P_B(X)| P_A(X)) + I(P_A(X)| P_B(X))$
\medskip

\noindent is a distance. 
Normalizing we obtain
\medskip

\hfil$d_N(P_A(X),P_B(X)) = {\displaystyle \frac{d(P_A(X),P_B(X))}
   {I(P_A(X)\cap P_B(X))}}\,$,\hfil
\medskip

\noindent with values in $[0,\!1]$. 

So the selected attribute will be that one 
that minimizes the measure: 
$d_N(P_{\cal C}(X),P_A(X))\vspace*{2mm}$.
\medskip

\noindent{\it Branching Strategy}
\smallskip

\noindent Usual TDIDT algorithms consider a branch for each value of the selected 
attribute. This strategy is not feasible when the number of
values is big (or even infinite). In our case the greatest number of values
for an attribute is 45 ---the tag set size---
which is considerably big (this means that the branching factor could be 
45 at every level of the tree%
\footnote{In real cases the branching factor is much lower since not all 
tags appear always in all positions of the context.}). 
Some systems perform a previous recasting of the attributes in order
to have only binary-valued attributes and to deal with binary trees
\cite{Magerman96}. 
This can always be done but the resulting features lose their
intuition and direct interpretation, and explode in number. 
We have chosen a mixed approach which consist of splitting for all
values and afterwards joining the resulting subsets into groups
for which we have not enough statistical evidence of being different 
distributions. This statistical evidence is tested with a $\chi^2$ test at a 5\%
level of significance. 
In order to avoid zero probabilities the following smoothing is performed.
In a certain set of examples, 
the probability of a tag $t_i$ is estimated by 
\smallskip

\hfil$\hat p(t_i)={\frac {|t_i|+{\frac {1}{m}}} {n+1}}\,$,\hfil
\medskip

\noindent where $m$ is the number of possible tags and $n$ the number of
examples.

Additionally, all the subsets that don't imply a reduction in the
{\it classification error} are joined together in order to have a bigger set
of examples to be treated in the following step of the tree
construction.
The classification error of a certain node is simply: 
$1- \max_{1\leq i\leq m}\ (\hat p(t_i))\,$.
\smallskip

Experiments reported in \cite{Marquez95} show that in this way more compact and
predictive trees are obtained.
\medskip

\noindent{\it Pruning the Tree}
\smallskip

\noindent Decision trees that correctly classify all examples of the training
set are not always the most predictive ones.
This is due to the phenomenon known as {\it over-fitting}. It occurs
when the training set has a certain amount of misclassified
examples, which is obviously the case of our training corpus (see section 
\ref{seccio-corpus}).
If we force the learning algorithm to completely classify the examples
then the resulting trees would fit also the noisy examples.

The usual solutions to this problem are: 1) Prune the tree, either during the 
construction process \cite{Quinlan93} or afterwards \cite{Mingers89b}; 2) Smooth the 
conditional probability distributions using fresh corpus%
\footnote{Of course, this can be done only in the case of statistical 
decision trees.} \cite{Magerman96}.

Since another important requirement of our problem is to have small trees
we have implemented a post-pruning technique.
In a first step the tree is completely expanded
and afterwards it is pruned following a minimal cost--complexity 
criterion \cite{Breiman84}. Roughly speaking this is  
a process that iteratively cut those subtrees producing only marginal 
benefits in accuracy, obtaining smaller trees at each step. The
trees of this sequence are tested using a, comparatively small, fresh 
part of the training set in order to decide which is the one with the
highest degree of accuracy on new examples.
Experimental tests \cite{Marquez95} have shown that the pruning process 
reduces tree sizes at about 50\% and improves their accuracy in a 2--5\%.
\medskip

\noindent{\it An Example}
\smallskip

\noindent Finally, we present a real example of the simple acquired contextual 
constraints for the conflict IN--RB (preposition-adverb).
\medskip

\noindent\hfil\epsfbox{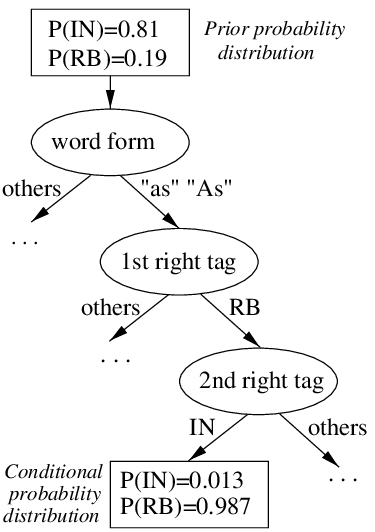}\hfil

\hfil \small{Figure~2: Example of a decision tree branch.}\hfil
\bigskip
\normalsize

The tree branch in figure~2 is translated into the following constraints: 
\smallskip

\indent\indent{\tt -5.81  <["as" "As"],IN> ([RB]) ([IN]);}
\newline
\indent\indent{\tt 2.366  <["as" "As"],RB> ([RB]) ([IN]);}
\medskip

\noindent which express the compatibility (either positive or
negative) of the word--tag pair in angle brackets with the given context.
The compatibility value for each constraint is the mutual information 
between the tag and the context \cite{Cover91}. It is directly computed from
the probabilities in the tree.

\section{Tagging Algorithm}
\label{seccio-relax}

Usual tagging algorithms are either n--gram oriented --such as
Viterbi algorithm \cite{Viterbi67}-- or ad--hoc for every case when they must
deal with more complex information.
\smallskip

We use relaxation labelling as a tagging algorithm. 
Relaxation labelling is a generic name for a family of iterative
algorithms which perform function optimization, based on local
information.  See \cite{Torras89} for a summary.
Its most
remarkable feature is that it can deal with any kind of constraints,
thus the model can be improved by adding any constraints available and
it makes the tagging algorithm independent of the complexity of the
model. \smallskip

The algorithm has been applied to
part--of--speech tagging \cite{Padro96}, and to shallow parsing \cite{Voutilainen97}.
\smallskip

The algorithm is described as follows:
 
  Let \(V=\{v_1,v_2,\ldots,v_n\}\) be a set of variables (words).

  Let \(t_i=\{t_1^i,t_2^i,\ldots,t_{m_i}^i\}\) be the set of possible
  labels (POS tags) for variable \(v_i\).

  Let \(CS\) be a set of constraints between the labels of the variables.
  Each constraint \(C\in CS\) states a ``compatibility value'' \(C_r\) for 
  a combination of pairs variable--label. 
  Any number of variables may be involved in a constraint. 
  
The aim of the algorithm is to find a weighted labelling\footnote{A
    weighted labelling is a weight assignment for each label of each
    variable such that the weights for the labels of the same variable
    add up to one.}
such that ``global consistency'' is maximized. Maximizing ``global
consistency'' is defined as maximizing for all \(v_i\),
\(\sum_j p^i_j \times S_{ij} \), where \(p^i_j\) is the weight 
for label \(j\) in variable \(v_i\) and
\(S_{ij}\) the support received by the same combination. The support
for the pair variable--label expresses {\em how compatible} that
pair is with the labels of neighbouring variables, according to
the constraint set. It is a vector optimization and doesn't 
maximize {\em only} the sum of the supports of all variables. It finds
a weighted labelling such that any other choice wouldn't increase the 
support for {\em any} variable.

The support is defined as the sum of the influence of every constraint
on a label.
$$
S_{ij}={\displaystyle\sum_{r \in R_{ij}}{Inf(r)}} 
$$
where: \\
\(R_{ij}\) is the set of constraints on label \(j\) for variable 
\(i\), i.e. the constraints formed by any combination of
variable--label pairs that includes the pair \((v_i,t^i_j)\).\\
\(Inf(r) = C_r \times p^{r_1}_{k_1}(m) \times \ldots
\times p^{r_d}_{k_d}(m)\), is the product of
the current weights\footnote{\(p^{r}_{k}(m)\) is the weight assigned to label
\(k\) for variable \(r\) at time \(m\).} for the labels appearing in the constraint except
\((v_i,t^i_j)\) (representing {\em how
applicable} the constraint is in the current context) 
multiplied by \(C_r\) which is the constraint compatibility value
(stating {\em how compatible} the pair is with the context). 
\smallskip

  Briefly, what the algorithm does is:
\begin{enumerate}
\item Start with a random weight assignment\footnote{We use lexical
    probabilities as a starting point.}.
\item Compute the support value for each label of each variable.
\item Increase the weights of the labels more compatible with the
         context (support greater than \(0\)) and decrease those of
         the less compatible labels (support less than
         \(0\))\footnote{Negative values for support indicate  {\em incompatibility}.},
         using the updating function:
$$
p^i_j(m+1) = \frac{{\displaystyle p^i_j(m)
    \times(1+S_{ij})}}{{\displaystyle
    \sum_{k=1}^{k_i}{p^i_k(m)\times(1+S_{ik})}}}
$$
$
\;\;\;\;\;\;\;\;\;\;\;\;\;\;\;\;\;\;\;\;\;\; {\rm where} \;\; -1\le S_{ij}\le +1
$
\item If a stopping/convergence criterion\footnote{We use the
         criterion of stopping when there are no more changes, although
         more sophisticated heuristic procedures are also used to stop
         relaxation processes \cite{Eklundh78,Richards81}.} is satisfied,
         stop, otherwise go to step 2.
\end{enumerate}

The cost of the algorithm is proportional to the product of the number of 
words by the number of constraints. 

\section{Description of the corpus}
\label{seccio-corpus}

 We used the Wall Street Journal corpus to train and test the system.
We divided it in three parts: $1,100$ Kw were used as a training set,
$20$ Kw as a model--tuning set, and $50$ Kw as a test set.
 
 The tag set size is 45 tags. $36.4\%$ of the words in the corpus are
 ambiguous, and the ambiguity ratio is $2.44$ tags/word over the
 ambiguous words, $1.52$ overall.
\smallskip

 We used a lexicon derived from training corpora, that contains
all possible tags for a word, as well as their lexical probabilities.
For the words in test corpora not appearing in the train set, we stored all
possible tags, but no lexical probability (i.e. we assume uniform
distribution)\footnote{That is, we assumed a
morphological analyzer that provides all possible tags for unknown
words.}.

 The noise in the lexicon was filtered by manually checking the lexicon entries
for the most frequent 200 words in the corpus\footnote{The 200 most frequent words
in the corpus cover over half of it.} to eliminate the tags due to
errors in the training set.
For instance the original lexicon entry 
(numbers indicate frequencies in the training corpus)
for the very common word {\it the} was
\smallskip

\noindent\hfil{\tt the CD 1 DT 47715 JJ 7 NN 1 NNP 6 VBP 1}\hfil
\smallskip

\noindent since it appears in the corpus 
with the six different tags: CD (cardinal), DT (determiner), JJ
(adjective), NN (noun), NNP (proper noun) and VBP (verb-personal
form). 
It is obvious that the only correct reading for {\it the} is
determiner.

 The training set was used to estimate bi/trigram statistics and to
 perform the constraint learning.

 The model--tuning set was used to tune the algorithm
parameterizations, and to write the linguistic part of the model. 
\smallskip

 The resulting models were tested in the fresh test set.

\section{Experiments and results}
\label{seccio-experiments}

The whole WSJ corpus contains 241 different classes of ambiguity.
The 40 most representative classes\footnote{In
terms of number of examples.} were selected for acquiring the
corresponding decision trees.
That produced 40 trees totaling up to 2995 leaf nodes, and covering 
83.95\% of the ambiguous words. Given that
each tree branch produces as many constraints as tags its leaf
involves, these trees were translated into 8473 context constraints.

We also extracted the 1404 bigram restrictions and the 17387 trigram
restrictions appearing in the training corpus.

Finally, the model--tuning set was tagged using a bigram model. The most
common errors commited by the bigram tagger were selected for
manually writing the sample linguistic part of the model, consisting
of a set of 20 hand-written constraints.

From now on {\bf C} will stands for the set of acquired context constraints, 
{\bf B} for the bigram model, {\bf T} for the trigram model, and 
{\bf H} for the hand-written constraints.
Any combination of these letters will indicate the joining of the 
corresponding models ({\bf BT}, {\bf BC}, {\bf BTC}, etc.). 

In addition, {\bf ML} indicates a baseline model containing no constraints
(this will result in a most-likely tagger) and {\bf HMM} stands for a
hidden Markov model bigram tagger \cite{Elworthy92}.

We tested the tagger on the 50 Kw test set using all the
combinations of the language models. Results are reported below.
\smallskip

The effect of the acquired rules on the number of errors for some of 
the most common cases is shown in table \ref{taula-separats}.
XX/YY stands for an error consisting of a word tagged YY when it
should have been XX. Table \ref{taula-tags} contains the meaning 
of all the involved tags.

{\small
\begin{table*}[htb] \centering
\begin{tabular}{|l|r||r||r|r||r|r||r|r|} \hline
                 &ML      &C     &B      &BC     &T      &TC    &BT     &BTC   \\ \hline
JJ/NN+NN/JJ      &73+137  &70+94 &73+112 &69+102 &57+103 &61+95 &67+101 &62+93 \\ \hline
VBD/VBN+VBN/VBD  &176+190 &71+66 &88+69  &63+56  &56+57  &55+57 &65+60  &59+61 \\ \hline
IN/RB+RB/IN      &31+132  &40+69 &66+107 &43+17  &77+68  &47+67 &65+98  &46+83 \\ \hline
VB/VBP+VBP/VB    &128+147 &30+26 &49+43  &32+27  &31+32  &32+18 &28+32  &28+32 \\ \hline 
NN/NNP+NNP/NN    &70+11   &44+12 &72+17  &45+16  &69+27  &50+18 &71+20  &62+15 \\ \hline
NNP/NNPS+NNPS/NNP&45+14   &37+19 &45+13  &46+15  &54+12  &51+12 &53+14  &51+14 \\ \hline
``that''         &187     &53    &66     &45     &60     &40    &57     &45    \\ \hline \hline
Total            &1341    &631   &820    &630    &703    &603   &731    &651   \\ \hline
\end{tabular}
\caption{\small{Number of some common errors commited by each model}}
\label{taula-separats}
\end{table*}
}

\begin{table}[htb] \centering
\begin{tabular}{l l} \hline
 NN       & Noun \\
 JJ       & Adjective\\ 
 VBD      & Verb -- past tense\\ 
 VBN      & Verb -- past participle\\
 RB       & Adverb \\ 
 IN       & Preposition \\
 VB       & Verb -- base form\\ 
 VBP      & Verb -- personal form\\ 
 NNP      & Proper noun \\ 
 NNPS     & Plural proper noun \\ \hline
\end{tabular}
\caption{\small{Tag meanings}}
\label{taula-tags}
\end{table}

 Figures in table \ref{taula-separats} show that in all cases the
 learned constraints led to an improvement.

 It is remarkable that when using {\bf C} alone, the
 number of errors is lower than with any bigram and/or trigram model, 
 that is, the acquired model performs better than the others estimated
 from the same training corpus.

 We also find that the cooperation of a bigram or trigram model with 
 the acquired one, produces even better results.
 This is not true in the cooperation of bigrams and trigrams with acquired
 constraints ({\bf BTC}), in this case the synergy is not enough to get a 
 better joint result. This might be due to the fact that the noise in {\bf B}
 and {\bf T} adds up and overwhelms the context constraints.
\smallskip

The results obtained by the baseline taggers can be found in table 
\ref{taula-resultats-base}
and the results obtained using all the learned constraints together 
with the bi/trigram models in table \ref{taula-resultats-our}.

\begin{table}[htb] \centering
\begin{tabular}{|l|r|r|} \hline
     &ambiguous &overall    \\ \hline \hline
ML   &$85.31\%$ &$94.66\%$  \\ \hline
HMM  &$91.75\%$ &$97.00\%$  \\ \hline
\end{tabular}
\caption{\small{Results of the baseline taggers}}
\label{taula-resultats-base}
\end{table}
\medskip
\begin{table}[htb] \centering
\begin{tabular}{|l|r|r|} \hline
     &ambiguous &overall    \\ \hline \hline
B    &$91.35\%$ &$96.86\%$  \\ \hline
T    &$91.82\%$ &$97.03\%$  \\ \hline
BT   &$91.92\%$ &$97.06\%$  \\ \hline \hline
C    &$91.96\%$ &$97.08\%$  \\ \hline
BC   &$92.72\%$ &$97.36\%$  \\ \hline
TC   &$92.82\%$ &$97.39\%$  \\ \hline
BTC  &$92.55\%$ &$97.29\%$  \\ \hline
\end{tabular}
\caption{\small{Results of our tagger using every combination of constraint kinds}}
\label{taula-resultats-our}
\end{table}

On the one hand, the results in tables \ref{taula-resultats-base} 
and \ref{taula-resultats-our} show that our tagger
performs slightly worse than a HMM tagger in the same conditions%
\footnote{Hand analysis of the errors commited by the algorithm
suggest that the worse results may be due to noise in 
the training and test corpora, i.e., relaxation algorithm seems to be 
more noise--sensitive than a Markov model. Further research is
required on this point.}, that is, when using only bigram information.

On the other hand, those results also show that since our tagger is 
more flexible than a HMM, it can easily accept more complex information to 
improve its results up to $97.39\%$ without modifying the algorithm.

\begin{table}[ht] \centering
\begin{tabular}{|l|r|r|} \hline
      &ambiguous &overall    \\ \hline \hline
H     &$86.41\%$ &$95.06\%$  \\ \hline
BH    &$91.88\%$ &$97.05\%$  \\ \hline
TH    &$92.04\%$ &$97.11\%$  \\ \hline
BTH   &$92.32\%$ &$97.21\%$  \\ \hline \hline
CH    &$91.97\%$ &$97.08\%$  \\ \hline
BCH   &$92.76\%$ &$97.37\%$  \\ \hline
TCH   &$92.98\%$ &$97.45\%$  \\ \hline
BTCH  &$92.71\%$ &$97.35\%$  \\ \hline
\end{tabular}
\caption{\small{Results of our tagger using every combination of constraint kinds and hand written constraints}}
\label{taula-resultats-manuals}
\end{table}

 Table \ref{taula-resultats-manuals} shows the results adding the hand
 written constraints. 
 The hand written set is very small and only covers
 a few common error cases. That produces poor results when using them
 alone (H), but they are good enough to raise the results given by the
 automatically acquired models up to $97.45\%$.

   Although the improvement obtained might seem small, it must be taken
  into account that we are moving very close to the best achievable
  result with these techniques.

  First, some ambiguities can only be solved with semantic
  information, such as the Noun--Adjective ambiguity for word
  {\em principal} in the phrase {\em the principal office}. It could be an
  adjective, meaning {\em the main office}, or a noun,
  meaning {\em the school head office}. 

  Second, the WSJ corpus contains noise (mistagged words) that affects
  both the training and the test sets. The noise in the training set
  produces noisy --and so less precise-- models. In the test set, it 
  produces a wrong estimation of accuracy, since correct answers are
  computed as wrong and vice-versa.

  For instance, verb participle
  forms are sometimes tagged as such ({\em VBN}) and also as adjectives ({\em JJ})
  in other sentences with no structural differences:
\begin{itemize}
\item {\tt... failing\_VBG to\_TO voluntarily\_RB submit\_VB 
the\_DT {\em requested\_VBN} information\_NN ...}

\item {\tt... a\_DT large\_JJ sample\_NN of\_IN {\em married\_JJ} women\_NNS 
with\_IN at\_IN least\_JJS one\_CD child\_NN ...}
\end{itemize}

   Another structure not coherently tagged are noun
   chains when the nouns are ambiguous and can be also adjectives:
\begin{itemize}

\item {\tt ... Mr.\_NNP Hahn\_NNP ,\_, the\_DT 62-year-old\_JJ chairman\_NN 
and\_CC {\em chief\_NN executive\_JJ officer\_NN} of\_IN 
Georgia-Pacific\_NNP Corp.\_NNP ...}

\item {\tt... Burger\_NNP King\_NNP 's\_POS {\em chief\_JJ executive\_NN officer\_NN} ,\_, 
Barry\_NNP Gibbons\_NNP ,\_, stars\_VBZ in\_IN ads\_NNS saying\_VBG ...}

\item {\tt... and\_CC Barrett\_NNP B.\_NNP Weekes\_NNP ,\_, chairman\_NN ,\_, 
president\_NN and\_CC {\em chief\_JJ executive\_JJ officer\_NN} .\_. }

\item {\tt... the\_DT company\_NN includes\_VBZ Neil\_NNP Davenport\_NNP ,\_, 
47\_CD ,\_, president\_NN and\_CC {\em chief\_NN executive\_NN officer\_NN} ;\_:}
\end{itemize}

  All this makes that the performance cannot reach $100\%$, and that
  an accurate analysis of the noise in WSJ corpus should be
  performed to estimate the actual upper bound that a tagger can
  achieve on these data. This issue will be addressed in further work.

\section{Conclusions}

 We have presented an automatic constraint learning algorithm
based on statistical decision trees.

 We have used the acquired constraints in a part--of--speech tagger that
allows combining any kind of constraints in the language model. 

 The results obtained show a clear improvement in the performance  
when the automatically acquired constraints are added to the model.
That indicates that relaxation labelling is a flexible algorithm able
to combine properly different information kinds, and that the constraints
acquired by the learning algorithm capture relevant context
information that was not included in the n--gram models. 

 It is difficult to compare the results to other works, since the accuracy 
 varies greatly depending on the corpus, the tag set, and the lexicon
 or morphological analyzer used. The more similar conditions reported in
 previous work are those experiments performed on the WSJ corpus:
 \cite{Brill92} reports $3$-$4\%$ error rate, and \cite{Daelemans96}
 report $96.7\%$ accuracy. We obtained a $97.39\%$ accuracy with trigrams plus
 automatically acquired constraints, and $97.45\%$ when hand written
 constraints were added. 

\section{Further Work}

 Further work is still to be done in the following directions:
\begin{itemize}
\item Perform a thorough analysis of the noise in the WSJ corpus to
determine a realistic upper bound for the performance that can be
expected from a POS tagger. 
\end{itemize}

\noindent On the constraint learning algorithm:
\begin{itemize}
\item Consider more complex context features, such as non--limited
  distance or barrier rules in the style of \cite{Samuelsson96}.
\item Take into account morphological, semantic and other kinds of information.
\item Perform a global smoothing to deal with low--frequency ambiguity classes.
\end{itemize}

\noindent On the tagging algorithms
\begin{itemize}
\item Study the convergence properties of the algorithm to decide
  whether the lower results at convergence are produced by the noise in the corpus.
\item Use back-off techniques to minimize interferences between
  statistical and learned constraints.
\item Use the algorithm to perform simultaneously POS tagging and word sense 
disambiguation, to take advantage of cross influences between both kinds of 
information.
\end{itemize}


\begin{thebibliography}{fullname}

\bibitem[\protect\citename{Aha \bgroup et al.\egroup}1991]{Aha91} 
                 D.W.~Aha, D.~Kibler and M.~Albert.
\newblock 1991
\newblock Instance--based learning algorithms.
\newblock In {\em Machine Learning.} 7:37-66.
Belmont, California.

\bibitem[\protect\citename{Breiman \bgroup et al.\egroup}1984]{Breiman84} 
                 L.~Breiman, J.H.~Friedman, R.A.~Olshen and C.J.~Stone.
\newblock 1984
\newblock Classification and Regression Trees.
\newblock The Wadsworth Statistics/Probability Series. Wadsworth International Group, 
Belmont, California.

\bibitem[\protect\citename{Brill}1992]{Brill92} E.~Brill.
\newblock 1992
\newblock A Simple Rule-Based Part-of-Speech.
\newblock In {\em Proceedings of the Third Conference on Applied Natural Language Processing.} ACL. 

\bibitem[\protect\citename{Brill}1995]{Brill95} E.~Brill.
\newblock 1995
\newblock Unsupervised Learning of Disambiguation Rules for Part--of--speech Tagging.
\newblock In {\em Proceedings of 3rd Workshop on Very Large Corpora.} Massachusetts.

\bibitem[\protect\citename{Cover and Thomas}1991]{Cover91} T.M.~Cover
                and J.A.~Thomas (Editors)
\newblock 1991
\newblock Elements of information theory.
\newblock John Wiley \& Sons.

\bibitem[\protect\citename{Cutting  \bgroup et al.\egroup }1992]{Cutting92}
              D.~Cutting,  J.~Kupiec, J.~Pederson and P.~Sibun.
\newblock 1992
\newblock A Practical Part--of--Speech Tagger.
\newblock In {\em Proceedings of the Third Conference on Applied Natural Language Processing.}, ACL. 

\bibitem[\protect\citename{Eklundh and Rosenfeld}1978]{Eklundh78}
                J.~Eklundh and A.~Rosenfeld.
\newblock 1978
\newblock Convergence Properties of Relaxation Labelling.
\newblock Technical Report no. 701. 
\newblock Computer Science Center. University of Maryland.

\bibitem[\protect\citename{Elworthy}1992]{Elworthy92} 
               D.~Elworthy.
\newblock 1993
\newblock Part--of--Speech and Phrasal Tagging.
\newblock Technical report, SPRIT BRA--7315 Acquilex II, Working Paper WP \#10.

\bibitem[\protect\citename{Daelemans \bgroup et al.\egroup }1996]{Daelemans96}
 W.~Daelemans, J.~Zavrel, P.~Berck and S.~Gillis.
\newblock 1996
\newblock MTB: A Memory--Based Part--of--Speech Tagger Generator.
\newblock In {\em Proceedings of 4th Workshop on Very Large Corpora.} Copenhagen, Denmark.

\bibitem[\protect\citename{Garside \bgroup et al.\egroup }1987]{Garside87}
        R.~Garside, G.~Leech and G.~Sampson (Editors) 
\newblock 1987
\newblock {\em The Computational Analysis of English.} 
\newblock London and New York: Longman.

\bibitem[\protect\citename{Hindle}1989]{Hindle89} D.~Hindle.
\newblock 1989
\newblock Acquiring disambiguation rules from text.
\newblock In {\em Proceedings ACL'89}.

\bibitem[\protect\citename{Karlsson}1990]{Karlsson90} F.~Karlsson
\newblock 1990
\newblock Constraint Grammar as a Framework for Parsing Running Text. 
\newblock In H. Karlgren (ed.), {\it Papers presented to the 13th International Conference on Computational Linguistics, Vol. 3}. Helsinki. 168--173.

\bibitem[\protect\citename{Karlsson \bgroup et al.\egroup }1995]{Karlsson95} 
        F.~Karlsson, A.~Voutilainen, J.~Heikkil\"a and A.~Anttila. (Editors)
\newblock 1995
\newblock {\em Constraint Grammar: A Language--Independent System for
        Parsing Unrestricted Text.}
\newblock Mouton de Gruyter, Berlin and New York.

\bibitem[\protect\citename{L\'opez}1991]{Lopez91}
               R.~L\'opez.
\newblock 1991
\newblock A Distance--Based Attribute Selection Measure for Decision Tree Induction.
\newblock Machine Learning. Kluwer Academic.

\bibitem[\protect\citename{Losee}1994]{Losee94}
               R.M.~Losee.
\newblock 1994
\newblock Learning Syntactic Rules and Tags with Genetic Algorithms for Information
Retrieval and Filtering: An Empirical Basis for Grammatical Rules.
\newblock Information Processing \& Management, May.

\bibitem[\protect\citename{Magerman}1996]{Magerman96}
              M.~Magerman.
\newblock 1996
\newblock Learning Grammatical Structure Using Statistical Decision--Trees.
\newblock In {\em Lecture Notes in Artificial Intelligence
              1147. Grammatical Inference: Learning Syntax from Sentences.}
\newblock Proceedings ICGI-96. Springer.

\bibitem[\protect\citename{M\`arquez and
                Rodr\'{\i}guez}1995]{Marquez95} L.~M\`arquez and H.~Rodr\'{\i}guez.
\newblock 1995
\newblock Towards Learning a Constraint Grammar from Annotated Corpora Using Decision Trees.
\newblock ESPRIT BRA--7315 Acquilex II, Working Paper.

\bibitem[\protect\citename{McCarthy and Lehnert}1995]{McCarthy95}
J.F.~McCarthy and W.G.~Lehnert.
\newblock 1995
\newblock Using Decision Trees for Coreference Resolution.
\newblock In {\em Proceedings of 14th International Joint Conference on
          Artificial Intelligence (IJCAI'95).}

\bibitem[\protect\citename{Mingers}1989]{Mingers89a} J.~Mingers.
\newblock 1989
\newblock An Empirical Comparison of Selection Measures for
Decision--Tree Induction.
\newblock In {\em Machine Learning.} 3:319--342.

\bibitem[\protect\citename{Mingers}1989]{Mingers89b} J.~Mingers.
\newblock 1989
\newblock An Empirical Comparison of Pruning Methods for
Decision--Tree Induction.
\newblock In {\em Machine Learning.} 4:227--243.

\bibitem[\protect\citename{Padr\'o}1996]{Padro96} L.~Padr\'o.
\newblock 1996
\newblock POS Tagging Using Relaxation Labelling.
\newblock In {\em Proceedings of 16th International Conference on
          Computational Linguistics.} Copenhagen, Denmark.

\bibitem[\protect\citename{Quinlan}1986]{Quinlan86}
                J.R.~Quinlan.
\newblock 1986
\newblock Induction of Decision Trees.
\newblock In {\em Machine Learning.} 1:81--106.

\bibitem[\protect\citename{Quinlan}1993]{Quinlan93}
                J.R.~Quinlan.
\newblock 1993
\newblock C4.5: Programs for Machine Learning.
\newblock San Mateo, CA. Morgan Kaufmann.

\bibitem[\protect\citename{Richards \bgroup et al. \egroup
    }1981]{Richards81} J.~Richards, D.~Landgrebe and P.~Swain.
\newblock 1981
\newblock On the accuracy of pixel relaxation labelling.
\newblock {\em IEEE Transactions on System, Man and Cybernetics.} Vol. SMC--11

\bibitem[\protect\citename{Samuelsson \bgroup et al.\egroup }1996]{Samuelsson96} 
           C.~Samuelsson, P.~Tapanainen and A.~Voutilainen.
\newblock 1996
\newblock Inducing Constraint Grammars.
\newblock In {\em Proceedings of the 3rd International Colloquium on Grammatical Inference.}

\bibitem[\protect\citename{Schmid}1994]{Schmid94} H.~Schmid
\newblock 1994
\newblock Part--of--speech tagging with neural networks.  
\newblock In {\em Proceedings of 15th International Conference on
          Computational Linguistics.} Kyoto, Japan.

\bibitem[\protect\citename{Torras}1989]{Torras89} C.~Torras.
\newblock 1989
\newblock Relaxation and Neural Learning: Points of Convergence and Divergence.
\newblock {\em Journal of Parallel and Distributed Computing.} 6:217--244

\bibitem[\protect\citename{Viterbi}1967]{Viterbi67} A.J.~Viterbi.
\newblock 1967
\newblock Error bounds for convolutional codes and an asymptotically optimal decoding algorithm.
\newblock In {\em IEEE Transactions on Information Theory.} pg 260--269, April.

\bibitem[\protect\citename{Voutilainen and J\"arvinen}1995]{Voutilainen95}
  A.~Voutilainen and T.~J\"arvinen.
\newblock 1995
\newblock Specifying a shallow grammatical representation for parsing purposes.
\newblock  In {\em Proceedings of the 7th meeting of the European
  Association for Computational Linguistics.} 210--214.

\bibitem[\protect\citename{Voutilainen and Padr\'o}1997]{Voutilainen97}
  A.~Voutilainen and L.~Padr\'o.
\newblock 1997
\newblock Developing a Hybrid NP Parser.
\newblock In {\em Proceedings of ANLP'97.}

\end{thebibliography}
\end{document}